\pdfoutput=1

\documentclass[11pt]{article}

\usepackage[]{acl}

\usepackage{times}
\usepackage{latexsym}
\usepackage{algorithm}
\usepackage{algorithmic}
\usepackage{graphicx}
\usepackage{amsfonts}

\usepackage{multirow} 
\usepackage{booktabs} 
\usepackage{xcolor}  
\usepackage[normalem]{ulem} 
\usepackage{soul}
\newcommand\hly{\bgroup\markoverwith
  {\textcolor{orange!30}{\rule[-.5ex]{2pt}{2.5ex}}}\ULon}
\newcommand\hlg{\bgroup\markoverwith
  {\textcolor{green!30}{\rule[-.5ex]{2pt}{2.5ex}}}\ULon}
\newcommand\hlc{\bgroup\markoverwith
  {\textcolor{cyan!30}{\rule[-.5ex]{2pt}{2.5ex}}}\ULon}
\newcommand\hlr{\bgroup\markoverwith
  {\textcolor{red!30}{\rule[-.5ex]{2pt}{2.5ex}}}\ULon}
\newcommand\hlgy{\bgroup\markoverwith
  {\textcolor{violet!30}{\rule[-.5ex]{2pt}{2.5ex}}}\ULon}
\newcommand\hlb{\bgroup\markoverwith
  {\textcolor{blue!30}{\rule[-.5ex]{2pt}{2.5ex}}}\ULon}

\colorlet{soulyellow}{yellow!50}
\colorlet{soulorange}{orange!30}
\colorlet{soullime}{lime!50}
\colorlet{soulgreen}{green!20}
\colorlet{soulviolet}{violet!20}
\colorlet{soulblue}{blue!20}
\colorlet{soulred}{red!20}

\usepackage{xspace}
\newcommand{\method}{\textbf{\textsc{Pics}}\xspace}

\usepackage{etoolbox}

\usepackage[T1]{fontenc}

\usepackage[utf8]{inputenc}

\usepackage{microtype}

\title{Psychology-guided Controllable Story Generation}

\author{Yuqiang Xie \quad Yue Hu\footnotemark[2] \quad Yunpeng Li \quad Guanqun Bi \quad Luxi Xing \quad Wei Peng\\
          Institute of Information Engineering, Chinese Academy of Sciences, Beijing, China \\
          School of Cyber Security, University of Chinese Academy of Sciences, Beijing, China \\
          \texttt{\{xieyuqiang,huyue,liyunpeng,biguanqun,xingluxi,pengwei\}@iie.ac.cn} \\}

\begin{document}

\maketitle
\renewcommand{\thefootnote}{\fnsymbol{footnote}}

\begin{abstract}
\footnotetext[2]{Corresponding author.}
Controllable story generation is a challenging task in the field of NLP, which has attracted increasing research interest in recent years.
However, most existing works generate a whole story conditioned on the appointed keywords or emotions, ignoring the psychological changes of the protagonist\footnotemark[3]\footnotetext[3]{In this work, we define the protagonist as the most frequently occurring character in a story \cite{Morrow1985ProminentCA}.}.
Inspired by psychology theories, we introduce global psychological state chains, which include the needs and emotions of the protagonists, to help a story generation system create more controllable and well-planned stories.
In this paper, we propose a \textbf{P}sychology-gu\textbf{I}ded \textbf{C}ontrollable \textbf{S}tory Generation System (\method) to generate stories that adhere to the given leading context and desired psychological state chains for the protagonist.
Specifically, psychological state trackers are employed to memorize the protagonist's local psychological states to capture their inner temporal relationships.
In addition, psychological state planners are adopted to gain the protagonist's global psychological states for story planning.
Eventually, a psychology controller is designed to integrate the local and global psychological states into the story context representation for composing psychology-guided stories.
Automatic and manual evaluations demonstrate that \method outperforms baselines, and each part of \method shows effectiveness for writing stories with more consistent psychological changes.
\end{abstract}

\section{Introduction}
\label{1}

Controllable Story Generation (CSG) is an important task in natural language processing (NLP) \cite{Porteous2009ControllingNG,Peng2018TowardsCS,Alabdulkarim2021AutomaticSG}.
It has also become one of the test methods for progress in artificial intelligence (AI).
Most existing state-of-the-art works \cite{Kong2021StylizedSG,Rashkin2020PlotMachinesOG,Paul2021COINSDG,xu-etal-2020-megatron} generate a story conditioned by the appointed keywords or emotions, with the help of remarkable pre-trained language models (PLM), like GPT-2 \cite{Radford2019LanguageMA} and BART \cite{DBLP:conf/acl/LewisLGGMLSZ20}.
While most of these systems have been able to generate fluent stories, CSG still has many issues to be explored.

\begin{figure}[t]
\centering
\includegraphics[width=0.48\textwidth]{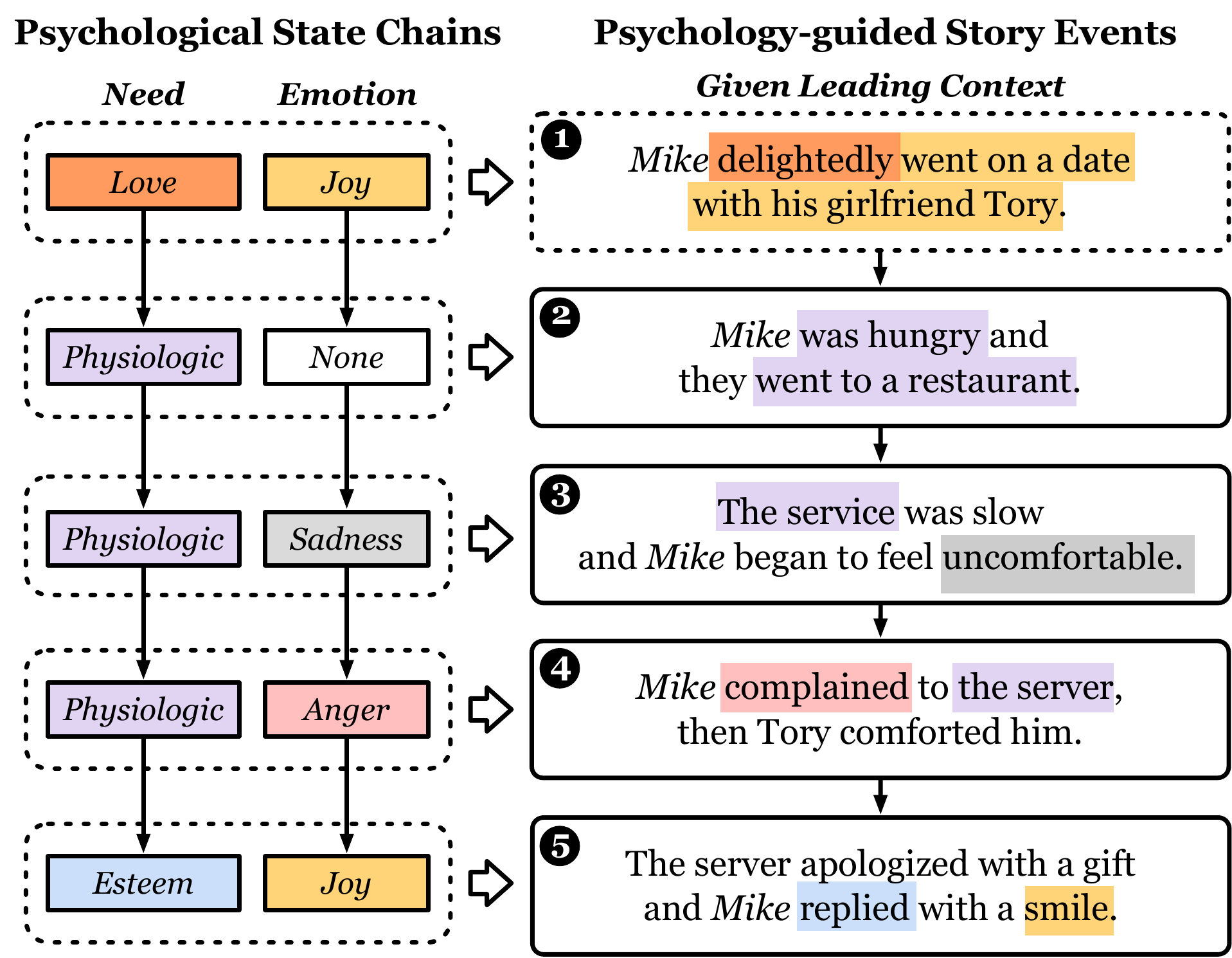} 
\caption{Example of psychology-guided controllable story generation conditioned on dotted frames (global psychological state chains, i.e., need/emotion, as well as leading context). Each psychological state and its corresponding tokens are highlighted in the same color.}
\label{fig-intro}
\end{figure}

In daily life, humans tend to create events driven by their needs (cause) and receive emotions (effect) after the events.
Similarly,  needs \cite{ricoeur1984time} and emotions \cite{Vonnegut1981PalmSA} play the central roles in creating reasonable stories in storytelling.
Recently, several works have begun developing CSG systems based on people’s expected emotional keywords or scores, such as \cite{DBLP:conf/emnlp/BrahmanC20} and \cite{Xu2020ControllableMP}.
Although these approaches can generate stories appointing the desired emotional signals, they are unable to control the storytelling as the protagonist's psychological state changes.
Another problem is that these methods only consider the current/previous emotions without global planning, which plays an important role in composing a story.

To address the aforementioned problems, we focus on taking the protagonist's global psychological state chains into account in controllable story generation.
Researches in cognitive psychology have shown that readers closely monitor the protagonist’s needs \cite{ricoeur1984time} and emotions \cite{Vonnegut1981PalmSA} while reading narratives.
At any point in a story, we represent the protagonist’s psychological state using multiple needs and emotions common in psychological theories.
\textit{Hierarchy of needs} of \citeauthor{Maslow2013ATO} has five categories (i.e., \textit{physiological need, stability, love and belonging, esteem} and \textit{self-actualization}) for describing human needs of a person.
\textit{Wheel of emotions} of \citeauthor{Plutchik1980AGP} proposes eight basic emotions (includes \textit{joy, trust, anger, surprise, sadness, disgust, fear} and \textit{anticipation}) to adequately portray a person.
Motivated by this, we define the psychological state chains as a sequence of five human needs and eight basic emotions that describe psychological states of a protagonist.

Given the protagonist's name and psychological state chains as well as the leading context, our goal is to generate a story about the leading context that adheres to the protagonist's psychological state chains.
As exemplified in Figure \ref{fig-intro}, the protagonist (\textit{Mike}) takes part in each story event controlled by given psychological state chains.
Note that, \textit{none} represents that \textit{Mike} has no need or emotion.
Each psychological state and its corresponding tokens are in the same color.
For example, in the second story event, \textit{Mike was hungry} and \textit{went to a restaurant} obviously embody the \textbf{physiologic need} of \textit{Mike}. Another example, as shown in the fourth story event, \textit{complain} action reflects \textit{Mike}'s \textbf{anger emotion}.
From a global perspective, intuitively, the anterior and hereafter psychological states separately provide the background and guidance for composing stories.
As illustrated in the third story event, anterior \textbf{physiologic need} leads to \textit{Mike's feeling uncomfortable} due to hungry, and hereafter \textbf{anger emotion} guides the setting of \textit{slow service} plot suspense.

To generate stories that adhere to the given leading context and the desired protagonist's global psychological state chains, we propose \method (\textbf{P}sychology-gu\textbf{I}ded \textbf{C}ontrollable \textbf{S}tory Generation System), a Transformer-based \cite{DBLP:conf/nips/VaswaniSPUJGKP17} architecture.
Specifically, psychological state trackers are employed to memorize the local psychological states for capturing temporal relationships among psychological states.
And, psychological state planners are adopted to gain the protagonist's global psychological states for planning the storytelling.
In the end, a psychology controller is designed to integrate the local and global psychological states into the story context representation for composing psychology-guided stories.
Based on the extracted data from publicly available \textit{Story Commonsense} \cite{DBLP:conf/acl/KnightCSRB18} dataset, experimental results demonstrate that \method outperforms baselines, and the psychological state trackers, planners as well as the psychology controller are important for generating stories with more consistent psychological changes.

\section{Related Work}

Early story generation systems relied on symbolic planning \cite{Prez2001MEXICAAC,Porteous2009ControllingNG,Riedl2010NarrativePB}, which had domain restriction and massive cost of feature engineering. Recent seq2seq storytelling models \cite{Roemmele2016WritingSW,Jain2017StoryGF} had partially alleviated these problems, most of which focused on learning better representation for a story \cite{Martin2018EventRF,Xu2018ASM,Fan2018HierarchicalNS,Fan2019StrategiesFS,Yao2019PlanAndWriteTB}.

To introduce semantic knowledge into story generation, many methods also employed large-scale pre-trained language models (LM) based on Transformer \cite{DBLP:conf/nips/VaswaniSPUJGKP17}, like GPT-2 \cite{Radford2019LanguageMA} and BART \cite{DBLP:conf/acl/LewisLGGMLSZ20}. After in-domain training, these models can generate fluent and coherent text, which can be used in story generation \cite{Qin2019CounterfactualSR,Guan2020AKP,clseg2022xie} and dialogue systems \cite{Budzianowski2019HelloIG,Wolf2019TransferTransfoAT}. However, they lacked the ability of controllable generation, such as expressing specific goals.

Further, aiming at controllable story generation, works had been introduced to control different attributes of the generated text, such as keyword \cite{Fan2018ControllableAS}, style \cite{Wang2017SteeringOS} and length \cite{Kikuchi2016ControllingOL}.
For example, \citeauthor{Tambwekar2019ControllableNS} \citeyear{Tambwekar2019ControllableNS} introduced reinforcement learning to generate a goal-driven storyline, which is a sequence of event tuples.
PPLM \cite{Dathathri2020PlugAP} used attribute classifiers to guide text generation without further training of LM.
PLOTMACHINES \cite{Rashkin2020PlotMachinesOG} transformed an outline into a coherent story by tracking the dynamic plot states.
\citeauthor{Kong2021StylizedSG} \citeyear{Kong2021StylizedSG} first planned the stylized keywords and then generated the whole story with the guidance of the keywords.
And many works considered commonsense knowledge as an attribute for CSG.
\citeauthor{Ammanabrolu2021AutomatedSV} \citeyear{Ammanabrolu2021AutomatedSV} performed story generation using soft causal relations, which automatically extracted from existing natural language plot summaries.
\citeauthor{Paul2021COINSDG} \citeyear{Paul2021COINSDG} used the contextualized commonsense inference rules generated by COMET \cite{DBLP:conf/acl/BosselutRSMCC19} based model to produce a coherent story ending.

Most related to this work, many methods generated text with a specific sentiment or emotion \cite{Zhou2018EmotionalCM,Huang2018AutomaticDG,Zhou2018MojiTalkGE,DBLP:conf/acl/SongZLXH19}.
\citeauthor{DBLP:conf/acl/KnightCSRB18} \citeyear{DBLP:conf/acl/KnightCSRB18} present an annotation framework specifically designed to examine the mental states of characters in commonsense based stories.
There are some limitations to incorporating sentiment, emotion or psychological state for story generation.
Previous work modeled characters but not sentiment \cite{DBLP:conf/naacl/ClarkJS18,DBLP:conf/aaai/LiuLYHL0020}.
\citeauthor{Peng2018TowardsCS} \citeyear{Peng2018TowardsCS} and \citeauthor{DBLP:conf/acl/LuoDYLCSS19} \citeyear{DBLP:conf/acl/LuoDYLCSS19} controlled the overall sentiment for story ending generation.
\citeauthor{DBLP:conf/conll/WeberSKBC20} \citeyear{DBLP:conf/conll/WeberSKBC20} incorporated sentiment trajectory by a new task that “filling in” a story.
\citeauthor{DBLP:conf/emnlp/BrahmanC20} \citeyear{DBLP:conf/emnlp/BrahmanC20} modeled the emotional trajectory of the protagonist for story generation.
\citeauthor{Xu2020ControllableMP} \citeyear{Xu2020ControllableMP} generated a story with multiple emotional changes of protagonists based on the given characters and the corresponding psychological state lines.
These works are limited to the guiding of emotion scores or tokens or/and target the ending sentence.
Lately, \citeauthor{xie2022comma} \citeyear{xie2022comma} modeling the relationship among motivations, actions and emotions based on human activities (i.e. story events), which lacks consideration of the global changes in a story.
Different from the above methods, we respectively model the local and global psychological state changes of the protagonist as the story progresses, which is more central to storytelling than the emotion trajectory.

\section{Task Definition}
We formulate our psychology-oriented controllable story generation task in the following. Note that, the length of the whole story is $5$ in this paper, and the output story event is in the $m$-th time point. Table \ref{tab:example-task-def} shows an example of our task.
\paragraph{Input}
The context $\mathbb{X}=(\mathcal{X}_1,\mathcal{X}_2,\ldots,\mathcal{X}_{m-1})$ to the current story event with $m-1$ events, where the $i$-th event $\mathcal{X}_i=(x^1_i,x^2_i,\ldots,x^k_i)$ consists of $k$ words. The name of protagonist $\mathbb{P}=(p_1,p_2,\ldots,p_m)$ to indicate the expected participant of the generated story event, the elements are the same in $\mathbb{P}$ in our setting. The protagonist's global psychological state chains, including the need chain $\mathbb{A_N}=(n_1,n_2,\ldots,n_5)$ and the emotion chain $\mathbb{A_E}=(e_1,e_2,\ldots,e_5)$. The protagonist's local psychological states from $\mathbb{A_N}$ and $\mathbb{A_E}$, including the need history $\mathbb{N}=(n_1,n_2,\ldots,n_m)$ and the emotion history $\mathbb{E}=(e_1,e_2,\ldots,e_m)$. $n_{i}$ and $e_{i}$ represent the protagonist's need and emotion for the $i$-th story event, where $i \in [1,m]$.

\paragraph{Output}
$\mathcal{Y}_m=(y_1,y_2,\ldots,y_r)$ (also $X_{m}$ in the next time step) stands for the current story event that consists of $r$ words, based on the protagonist's name $\mathbb{P}$, the need history $\mathbb{N}$, the emotion history $\mathbb{E}$, the need chain $\mathbb{A_N}$ and the emotion chain $\mathbb{A_E}$, where $y_i$ is the $i$-th word.

\section{Methodology}
\label{sec4}
\begin{figure*}[t]
  \centering
  \includegraphics[width=0.9\textwidth]{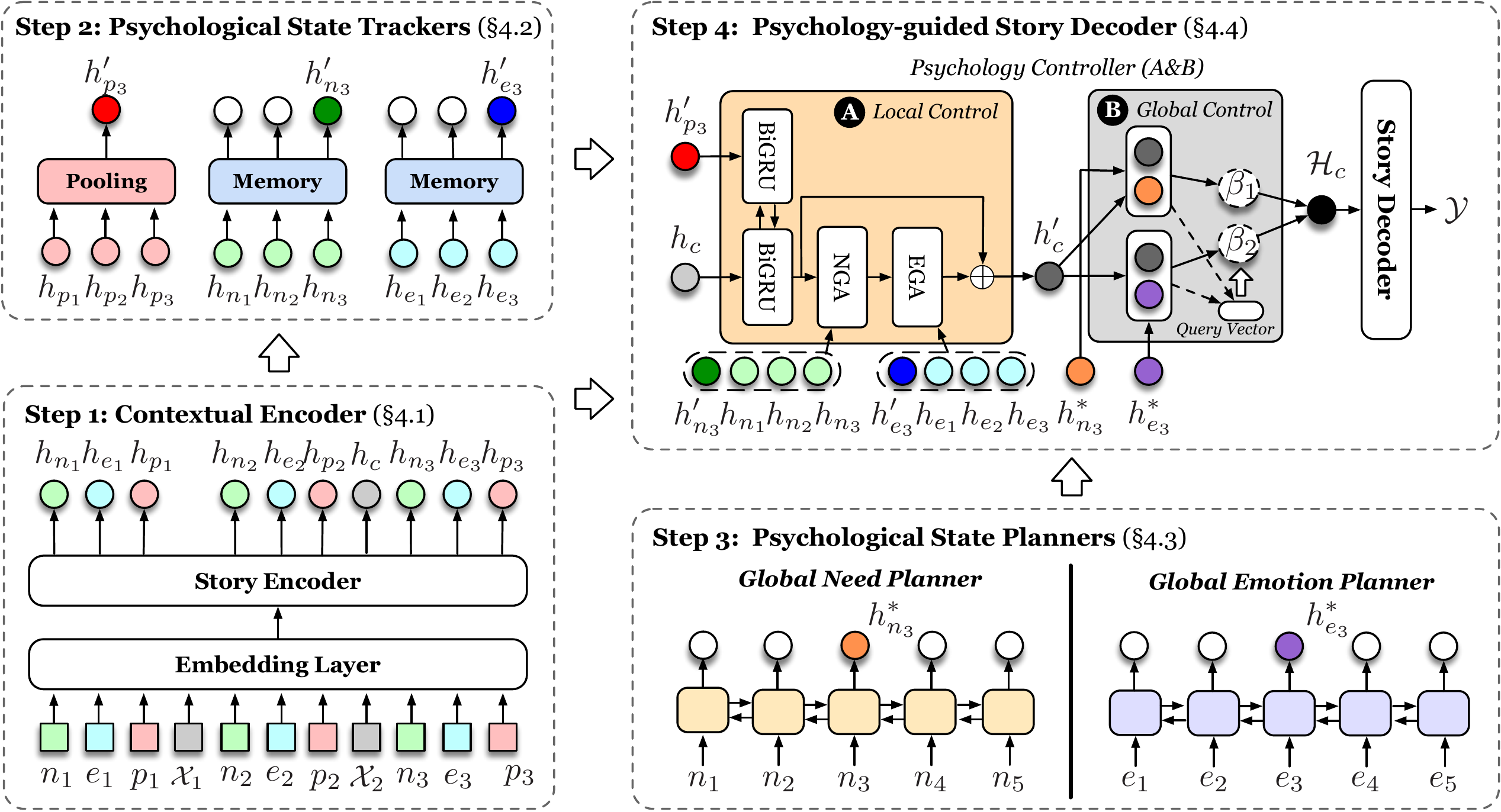} 
  \caption{Overview of \method with time point m=3 \textbf{(Step 1-4)}.
    In step 1, contextual encoder converts input into contextual representation (§ \ref{sec4-1}).
    In step 2, we design psychological state trackers to capture temporal relations for the protagonist’s character information, local need and emotion (§ \ref{sec4-2}).
    In step 3, two psychological state planners output the global psychological state through modeling the completed need and emotion chains (§ \ref{sec4-3}).
    In step 4, conditioned on the protagonist's local and global psychological states, the decoder generates psychology-guided stories with a psychology controller (A\&B) (§ \ref{sec4-4}).}
  \label{model-fig}
\end{figure*}

The overall architecture of our proposed \method system is illustrated in Figure \ref{model-fig}.
In the following, we will describe each component in more detail.

\begin{table}[t]
    \centering
    \scalebox{0.65}{
    \begin{tabular}{r|l}
    \toprule
       \bf Protagonist  &  Donald, He, Donald, He, He \\
       \bf Need Chain  &  esteem, esteem, esteem, esteem, esteem \\
       \bf Emotion Chain  &  joy, sadness, sadness, sadness, joy \\
       \bf Leading Context   & Donald was a senator. \\
       \midrule
       \bf Event $\mathcal{Y}_2$ &  \multicolumn{1}{p{8cm}}{He ran as an indie candidate.} \\
       \bf Event $\mathcal{Y}_3$ &  \multicolumn{1}{p{8cm}}{Donald wanted better implementation of his policies.} \\
       \bf Event $\mathcal{Y}_4$ &  \multicolumn{1}{p{8cm}}{He decided to run in the next term as a Republican.} \\
       \bf Event $\mathcal{Y}_5$ &  \multicolumn{1}{p{8cm}}{He won again, and is now in a better position.} \\
   \bottomrule
    \end{tabular}}
    \caption{An example of our task.}
    \label{tab:example-task-def}
\end{table}

\subsection{Contextual Encoder (Step 1)}
\label{sec4-1}

In order to capture the contextual semantic information for the story context and the protagonist’s historical psychological states, we reconstruct the input of the embedding layer in the backbone BART \cite{DBLP:conf/acl/LewisLGGMLSZ20} model.

\begin{figure}[ht]
  \centering
  \includegraphics[width=0.45\textwidth]{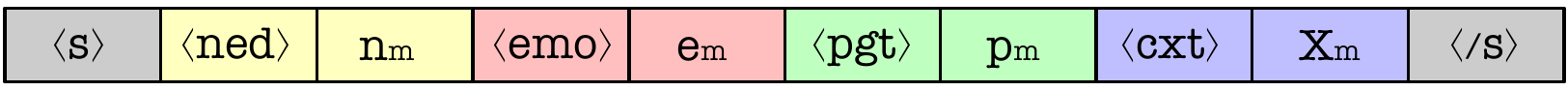}
  \caption{Reconstructed input of the word embedding layer in BART for $m$-th event.}
  \label{fig-embedding}
\end{figure}

\begin{figure}[t]
  \centering
  \includegraphics[width=0.3\textwidth]{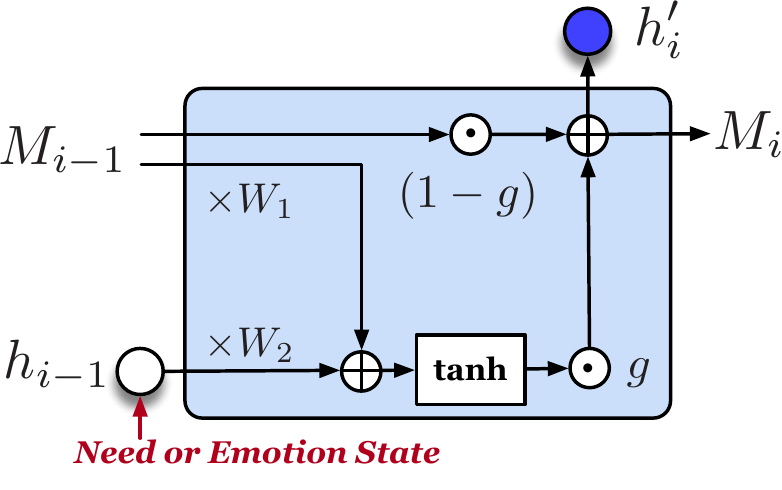} 
  \caption{Details of Memory Units in Psychological State Trackers for need and emotion states.}
  \label{model-memory}
\end{figure}

As illustrated in Figure \ref{fig-embedding}, we employ new special $\{\langle \texttt{ned} \rangle, \langle \texttt{emo} \rangle, \langle \texttt{pgt} \rangle \}$ tokens to delimit each protagonist's need, emotion and name grounded in the story context.
In addition, we utilize a special $\langle \texttt{cxt} \rangle$ token to delimit each story event of the context:
\begin{equation}
  b_{m} = \texttt{Emb}(n_1, e_1, p_1, X_1,\ldots,n_{m},e_{m},p_{m})
  \label{eq1}
\end{equation}
Then, we acquire representations of $i$-th psychological states $h_{n_i},h_{e_i},h_{p_i}$ by extracting the hidden states of special $\{\langle \texttt{ned} \rangle, \langle \texttt{emo} \rangle, \langle \texttt{pgt} \rangle \}$ tokens on the top of encoder.

Similarly, story context representation is corresponding to the hidden state of ($m-1$)-th special $\langle \texttt{cxt} \rangle$ token.

\subsection{Psychological State Trackers (Step 2)}
\label{sec4-2}

For the purpose of remembering and updating the protagonist's psychological states that have been mentioned, we design psychological state trackers for the protagonist's character information\footnotemark[4]\footnotetext[4]{In this paper, we regard the representation of the protagonist's name as his/her character information.}, needs and emotions.

\paragraph{Protagonist's Character Information}
\label{sec4-2-1}
Based on several story events, humans can easily guess the protagonist's character information. We argue that the resulting representation can stand for the protagonist's character information via pooling hidden states of the protagonist's name which is grounded in story events.
\begin{eqnarray}
    h^{\prime}_{p} = \texttt{Pooling}(\{h_{p_i}\}^{m}_{i=1})
    \label{eq4}
\end{eqnarray}
In this work, \texttt{Pooling} is Mean-Pooling and is used as a tracker for $p_i$ to conclude the moderate representation of the protagonist's character information.

\paragraph{Protagonist's Needs}
To remember and update the mentioned protagonist's needs and emotions, we design trackers (memory block in Figure \ref{model-fig}) as follows:
\begin{equation}
    h^{\prime}_{n} = \texttt{Memory}(\mathbb{N})
    \label{eq5}
\end{equation}
As shown in Figure \ref{model-memory}, for memorizing mentioned needs, the memory unit $M_{n_{i}}$ is updated using $h_{n_{i-1}}$ and $M_{n_{i-1}}$, the output contextual needs representation:
\begin{equation}
    \widehat{h}_{n_{i-1}} = \texttt{tanh}(W_1 M_{n_{i-1}}+W_2 h_{n_{i-1}})
    \label{eq6}
\end{equation}
Futher, we use a gating mechanism, $g$, to allow the model to learn to flexibly control how much each cell in memory is updated, as below:
\begin{equation}
    g_{n_i} = \texttt{sigmoid}(W_3 M_{n_{i-1}}+W_4 {h}_{n_{i-1}})
    \label{eq7}
\end{equation}
\begin{equation}
    M_{n_{i}} = g_{n_i} \widehat{h}_{n_{i-1}}+(1-g_{n_i})M_{n_{i-1}}
    \label{eq8}
\end{equation}
\begin{equation}
    h^{\prime}_{n_m} = M_{n_{m}}
    \label{eq9}
\end{equation}
where $M_{n_{0}}$ is randomly initialized and all $W_*$ are trainable parameters.
\paragraph{Protagonist's Emotions}
In the same way with need, for historical emotions:
\begin{equation}
    h^{\prime}_{e_m} = \texttt{Memory}(\mathbb{E}) = M_{e_{m}}
    \label{eq10}
\end{equation}

In summary, we can obtain sequential psychological state changes through the above operations.

\subsection{Psychological State Planner (Step 3)}
\label{sec4-3}

Different from psychological state trackers, we obtain the global psychological states by encoding the global needs chain and emotions chain for story planning. As shown in Figure \ref{model-fig} (step 3), psychological state trackers use a \texttt{BiGRU} \cite{Cho2014OnTP} architecture.

\paragraph{Global Need Planner}
In the global need planner, the global need representations:
\begin{equation}
    [t_n^1,t_n^2,\ldots,t_n^5] = \texttt{BiGRU}(\mathbb{A_N})
    \label{eq11}
\end{equation}
\begin{equation}
    h^{*}_{n_j} = t_n^j
    \label{eq12}
\end{equation}
where $j$ represents the $j$-th need for generate next story event and $j$-th need in the global need chain.

\paragraph{Global Emotion Planner}
Similarly, computing global emotion representations are as below:
\begin{equation}
    h^{*}_{e_j} = \texttt{BiGRU}(\mathbb{A_E})[j] = t_e^i
    \label{eq13}
\end{equation}
note that, $\mathbb{A_N}$ and $\mathbb{A_E}$ are initialized by GloVe \cite{Pennington2014GloVeGV} embedding.

After obtaining global need and emotion representations, we feed them into the following step as a planning signal for guiding the story generation.

\subsection{Psychology-guided Decoder (Step 4)}
\label{sec4-4}

Conditioned on the protagonist's local and global psychological states, the decoder generates a psychology-guided story event by the following modules.

\subsubsection{Psychology Controller}

In order to control the story generation by protagonist's psychological states, we respectively integrate local and global psychological states \cite{ijcai2022wei,DBLP:journals/corr/abs-2203-00819} into the story context representation.

\paragraph{Local Control}
With the goal to integrate local psychological control information into the representation of story context $h_c$, a psychology controller is used to compute the interaction between $h_c$ and local psychological states (including $h^{\prime}_{p}$, $h^{\prime}_{n}$, $h^{\prime}_{e}$). First, $h^{\prime}_{p}$ guides the model with the protagonist's character information for generating the next story event, which uses a \texttt{BiGRU} \cite{Cho2014OnTP} architecture:
\begin{equation}
    \widetilde{h_{c}} = \texttt{BiGRU}(h_c,h^{\prime}_{p})[0]
    \label{eq14}
\end{equation}
where we extract the hidden state $\widetilde{h_{c}}$ as the story context representation considering character information.
Then, we employ an attention mechanism to integrate local psychological states (need and emotion) into the story context representation.
Firstly, need guided attention \texttt{NGA} is defined as follows:
\begin{equation}
{h^n_{c}}
= \texttt{NGA}(\widetilde{h_{c}}, \{h_{n_i}\}^{m}_{i=1})
= \sum_{i=1}^{m} \alpha_i  h_{n_i}
\label{eq15}
\end{equation}
\begin{equation}
 \{\alpha_i\}^{m}_{i=1}  = \textit{softmax} (\{{\widetilde{h_{c}}h^{T}_{n_i}}/{\sqrt{dim_1}}\}^{m}_{i=1} )
\label{eq16}
\end{equation}
\begin{equation}
\widehat{h^n_{c}}  = \texttt{Fus.N}(h^{\prime}_{n},{h^n_{c}})=\texttt{MLP}([h^{\prime}_{n},{h^n_{c}}])
\label{eq17}
\end{equation}
where $dim_1$ equals to the dimension of ${h^n_{c}}$. Similarly, \texttt{EGA} and \texttt{Fus.E} has the same operation as the above equations:
\begin{equation}
\widehat{h^e_{c}}  = \texttt{Fus.E}( \texttt{EGA}(\widehat{h^n_{c}}, \{h_{e_i}\}^{m}_{i=1}) )
\label{eq18}
\end{equation}
\begin{equation}
    h^{\prime}_{c} = \widehat{h^e_{c}}+\widetilde{h_{c}}
    \label{eq19}
\end{equation}

\paragraph{Global Control}
Aiming at further control composing stories with the global planning signal, this part is designed to dynamically integrate global psychological states.
In specific, a query vector $q$ is introduced to fuse psychology-blended context representations and global psychological states by attention mechanism as below:
\begin{equation}
s_n, s_e = \frac {q {[h^{\prime}_{c},h^*_{n_i}]}^{T}}{\sqrt{dim_2}},  \frac {q {[h^{\prime}_{c},h^*_{e_i}]}^{T}}{\sqrt{dim_2}}
\label{eq20}
\end{equation}
\begin{equation}
\beta_1, \beta_2  =   \textit{softmax}( s_n, s_e)
\label{eq21}
\end{equation}
\begin{equation}
\mathcal{H}_{c} = \texttt{MLP} (\beta_1 [h^{\prime}_{c},h^*_{n_i}]+ \beta_2 [h^{\prime}_{c},h^*_{e_i}])
\label{eq22}
\end{equation}
where $q$ is the query and $[h^{\prime}_{c},h^*_{n_i}],[h^{\prime}_{c},h^*_{e_i}]$ are the keys for attention. $[\cdot]$ denotes the concatenation operation. $dim_2$ equals to the dimension of $[h^{\prime}_{c},h^*_{n_i}]$. So that the model can adaptively choose the most important global psychological state for generating well-planned stories.

\subsubsection{Story Decoder}
We employ a left-to-right BART decoder to generate a story conditioned upon all input elements. Each layer of the decoder additionally performs cross-attention over the concatenation of the final hidden layer of the BART encoder and $\mathcal{H}_{c}$.
\begin{equation}
    P (y_t|\mathbb{X},\mathbb{P},\mathbb{N},\mathbb{E},\mathbb{A_N},\mathbb{A_E},y_{<t}) = \textit{softmax}(W_s{s}_{t})
    \label{eq23}
\end{equation}
\begin{equation}
    {s}_{t} = \texttt{Dec}(y_{<t},\texttt{Enc}(b_{m}), \mathcal{H}_{c})
    \label{eq24}
\end{equation}
where $\texttt{Enc}(b_{m})$ is the final hidden layer of the BART encoder. 

\subsubsection{Training}
The training objective is to minimize the negative log-likelihood $L$ of the ground truth story event:
\begin{equation}
\mathcal{L}=-\sum_{t=1}^{r} \log P\left(y_{t} \mid \mathbb{X},\mathbb{P},\mathbb{N},\mathbb{E},\mathbb{A_N},\mathbb{A_E},y_{<t}\right)
\label{eq25}
\end{equation}
Aiming to obtain the completed story, we iteratively generate the $m$-th story events $\mathcal{Y}_m$ with the forecast generated $\mathcal{Y}_1, ..., \mathcal{Y}_{m-1}$.

\section{Experiment}

\begin{table*}[t]
  \centering
  \scalebox{0.9}{
    \begin{tabular}{l|cccccc|cc}
      \toprule
      \textbf{Models}  & \textbf{PPL} $\downarrow$   & \textbf{BLEU-1} $\uparrow$  & \textbf{BLEU-2} $\uparrow$
      & \textbf{Rouge-1} $\uparrow$  & \textbf{Rouge-2} $\uparrow$  & \textbf{Rouge-L} $\uparrow$
      &  \textbf{NC} $\uparrow$ & \textbf{EC} $\uparrow$ \\
      \midrule
      Fusion    &    25.68        &     19.83       &      2.89      &    5.81         &  1.66        &     8.64        &       0.19          & 0.16          \\
      Plan\&Write &    19.43        &     20.15       &      3.56      &    6.23         &  1.85        &     8.81        &       0.31          & 0.19          \\
      PPLM      &    -            &     18.42       &      4.39      &    8.73         &  2.13        &     9.29        &       0.45          & 0.41          \\
      GPT-2 FT     &   18.21         &     22.67       &      6.42      &    9.97        &  2.25        &     9.98        &       0.34          & 0.37          \\
      BART FT      &   17.85         &      21.84      &     6.03       &     9.32       &  2.51        &     9.56       &       0.36          & 0.32          \\
      \midrule
      \method &   \textbf{16.73}           &      \textbf{23.51}      &   \textbf{6.89}
      &     \textbf{12.43}      & \textbf{3.83}   &      \textbf{11.28}
      &  \textbf{0.64}         & \textbf{0.45}          \\
      \bottomrule
  \end{tabular}}
  \caption{The results of automatic evaluation on test set considering common-used metrics and the designed metric (NC/EC) to test the psychological state consistency of stories. $\downarrow$/$\uparrow$ indicates the lower/higher, the better.}
  \label{table-gen}
\end{table*}

\subsection{Data}
\label{data-pre}
We choose a Story Commonsense \cite{DBLP:conf/acl/KnightCSRB18} that has been annotated with a similar setting to us. Story Commonsense is a large-scale dataset as a resource for training and evaluating the mental state tracking of characters in short commonsense stories. This dataset contains over 300k low-level annotations for character motivations and emotional reactions. Story Commonsense was proposed for studying need/emotion tracking. Each sentence is annotated for all characters, and there are 3 crowd-workers voting for each need/emotion. If the characters have no need or emotion, the psychological state will be labeled `none'.

Based on our task definition, we extract the story with a protagonist (occurs in more than 4 sentences) and the corresponding need/emotion chains from Story Commonsense. Note that, we select the \textbf{Top-1} need/emotion label, based on annotators' voting scores to make up the psychological chains in our data set. If several labels have the best score of all, we will choose a low-level need label of Maslow's needs or a random emotion label. Following the 8:1:1 splitting ratio, we obtain 2,570/321/321 five-sentence stories for train/dev/test sets. In our psychology-guided CSG task, we generate a story event based on the story context, protagonist's name, need chain, and emotion chain. Therefore, each story will be reformed into 4 samples following our setting (i.e. 10,280/1283/1283 samples).

\begin{table}[ht]
  \centering
  \scalebox{0.9}{
    \begin{tabular}{cccc}
      \toprule[1pt]
      \multicolumn{2}{c}{\textbf{NC}} & \multicolumn{2}{c}{\textbf{EC}} \\
      \textbf{Accuracy}      & \textbf{F1-Score}     & \textbf{Accuracy}          & \textbf{F1-Score}         \\
      \toprule[0.5pt]
             64.6           & 64.8      &  56.8             & 59.2   \\
      \toprule[1pt]
  \end{tabular}}
  \caption{Accuracy and F1-Score of RoBERTa classifier for need/emotion on dev set, respectively.}
  \label{table-ncec}
\end{table}

\subsection{Implement Details}

For a fair comparison, we train our proposed models and the baselines with the same input (leading context and global need/emotion chains) that are automatically extracted from Story Commonsense \cite{DBLP:conf/acl/KnightCSRB18}. 
Our proposed models follow the setting of BART large \cite{DBLP:conf/acl/LewisLGGMLSZ20} model with 12 layers in each of the encoder and decoder and a hidden size of 1024.
The stories are encoded using BPE with a vocabulary size of 50,257.
We set the maximum sequence length to $100$ tokens, as it is large enough to contain all inputs.
We use Adam optimization with an initial learning rate of $0.00001$.
All models were trained until there was no improvement in the validation set performance.
During training, we use a label smoothed cross-entropy loss, with the smoothing parameter set to $0.1$.
At inference time, we set beam size as $5$, and remove duplicated trigrams in beam search.
We use the HuggingFace
\footnotemark[5]\footnotetext[5]{https://github.com/huggingface/transformers}~\cite{DBLP:conf/emnlp/WolfDSCDMCRLFDS20} PyTorch~\cite{DBLP:conf/nips/PaszkeGMLBCKLGA19} implementation\footnotemark[6]\footnotetext[6]{We will make our dataset and code publicly available at \url{https://github.com/IndexFziQ/PICS}.} on Tesla V100 GPU.

\subsection{Evaluation Metrics}
\paragraph{Automatic Metrics}
We use the following metrics for automatic evaluation:
\textbf{(1) Perplexity (PPL)} is an indicator of fluency. A smaller value is better.
\textbf{(2) BLEU} \cite{DBLP:conf/acl/PapineniRWZ02} is used for evaluating the overall quality of the generated story. We use n=1, 2.
\textbf{(3) Rouge} \cite{Lin2004ROUGEAP} with n=1, 2, L is used to measure the similarity between automatically generated and reference results.
\textbf{(4) Need/Emotion Consistency (NC/EC)} It is a learn-able automatic metric. We fine-tune a RoBERTa \cite{Liu2019RoBERTaAR} large model on the Story Commonsense \cite{DBLP:conf/acl/KnightCSRB18} train set as a classifier to distinguish whether a story event is corresponding to a \textbf{Top-1} need/emotion. Table \ref{table-ncec} shows results of NC/EC.

\paragraph{Manual Metrics}
We also conduct a manual evaluation of generated psychology-guided stories. Following \citet{DBLP:conf/acl/SongZLXH19}, crowd-workers are required to evaluate actions on a 0-3 scale (3 being very good) from two different perspectives: 

\textbf{(1) Content Quality} to indicate whether the generated story is fluent. \textbf{(2) Content Rationality} to assess whether it follows the given needs and emotions which is reasonable and consistent. 

During the manual evaluation, we display the input (leading context and global need/emotion arc) and two stories generated by the two models being compared. To avoid prejudice, we randomly changed the order in which the stories in the two models were displayed to the crowd-workers. We provided crowd-workers with instructions to explain the annotations and provided examples. Following this process, each pair of stories is annotated by three crowd-workers.

\subsection{Experimental Results}

\subsubsection{Baselines}

For a fair comparison, we train \method and the baselines with the same input (leading context and global need/emotion chains).

We compare our base storytelling model, \method, with following state-of-the-art models:
\begin{enumerate} 
    \item \textbf{Fusion}~\cite{Fan2018HierarchicalNS}, a storytelling model that first pre-trains a convolutional seq2seq model, then fixes the trained model and passes it to the second clone model with fusion mechanism.
    \item \textbf{Plan\&Write}~\cite{Yao2019PlanAndWriteTB}, another storytelling model first generates a plot as a sequence of keywords with the given leading context and then conditioned on the plot it generates the text of the story.
    \item \textbf{PPLM}~\cite{Dathathri2020PlugAP}, which can be extended to accept psychological state chains for controlling story generation. We use psychological state chains as the skeleton.
    \item \textbf{GPT-2 FT}~\cite{Radford2019LanguageMA} is a pre-trained generative LM. We use a medium-size version. We fine-tuned GPT-2 on our dataset following \cite{Guan2020AKP} with leading context and global need/emotion chains.
    \item \textbf{BART FT}~\cite{DBLP:conf/acl/LewisLGGMLSZ20} is a encoder-decoder architecture. We fine-tuned BART on our dataset following \cite{DBLP:conf/acl/LewisLGGMLSZ20} with leading context and global need/emotion chains.
\end{enumerate}
All of these models are trained, validated and tested on the same data splits described in \S\ref{data-pre}. In specific, we add emotion/need labels as additional input tokens to baseline models alongside the tokens for each story sentence. And, global emotion/need chains that concatenated with the story context are given to baseline models at each time step.

\begin{table}[t]
  \centering
  \scalebox{0.85}{
    \begin{tabular}{lcccc}
      \toprule[1pt]
      \multirow{2}{*}{\textbf{Model}} & \multicolumn{2}{c}{\textbf{Quality} $\uparrow$} & \multicolumn{2}{c}{\textbf{Rationality} $\uparrow$} \\
      & \textbf{Fluency}      & \textbf{Coherence}
       & \textbf{Need}          & \textbf{Emotion}     \\
      \toprule[0.5pt]
      PPLM     &      2.56       &  1.62             &  1.05             & 1.27 \\
      GPT-2 FT   &       \textbf{2.87}      &  1.58             &  1.73             & 1.82 \\
      BART FT   &      2.72       &   \textbf{1.79}            &  1.69             &  1.98      \\
      \toprule[0.5pt]
      \method   &      2.83       &   1.68            &  \textbf{2.16}            &  \textbf{2.34}        \\
      \toprule[1pt]
  \end{tabular}}
  \caption{Manual Evaluation in terms of content quality and content rationality about the generated stories.}
  \label{table-humaneval}
\end{table}

\begin{table}[t]
  \centering
  \scalebox{0.85}{
    \begin{tabular}{lcccc}
      \toprule[1pt]
      \textbf{Model}    & \textbf{AVG-B} $\uparrow$   & \textbf{AVG-R} $\uparrow$    &  \textbf{NC} $\uparrow$  & \textbf{EC} $\uparrow$  \\
      \toprule[0.5pt]
      \method &    15.89        &     8.65       &      0.64      &    0.45             \\
      w/o PST &  15.41          &     8.45       &      0.53      &    0.46               \\
      w/o PSP &  \bf\underline{15.24}          &     8.47       &      0.56      &    0.43                \\
      w/o PC &   15.34         &     \bf\underline{8.24}       &      0.55      &    0.42                \\
      w/o Need &   15.71         &     8.53       &      \bf\underline{0.42}      &    0.39                  \\
      w/o Emotion &   15.64         &     8.44       &      0.51      &    \bf\underline{0.35} \\
      \toprule[1pt]
  \end{tabular}}
  \caption{Ablation study of \method model and global need/emotion chains on dev set. PST: psychological state tracker. PSP: psychological state planner. PC: psychology controller.}
  \label{table-ablation}
\end{table}

\begin{table}[t] 
\centering
\scalebox{0.85}{
  \begin{tabular}{lcccc}
    \toprule[1pt]
    \method v.s.  & \textbf{Win}  & \textbf{Loss}   & \textbf{Tie} & $\kappa$   \\
    \midrule
    PPLM     &  \bf 54.6\%    &   18.5\%   &  26.9\%   &  30.2  \\
    GPT-2 FT     &  \bf 53.2\%    &   19.5\%   &  27.3\%   &  30.4  \\
    BART FT    &  \bf 52.3\%    &   18.4\%   &   29.3\%    &  27.9   \\
    \bottomrule[1pt]
    \end{tabular}}
\caption{Human A/B Test of \method. Results show that \method performs baseline models sufficiently.
$\kappa$ denotes Fleiss’ kappa (all are fair agreement or moderate agreement).
The p-value of scores < 0.05 in sign test.}
\label{table-abtest}
\end{table}

\subsubsection{Automatic Evaluation}
The results of the automatic evaluation are shown in Table \ref{table-gen}. Our model outperforms the variants of GPT-2 in terms of perplexity, and has higher BLEU and Rouge scores than all the baselines, indicating better fluency and more overlaps with the reference stories. Besides, in the view of NC and EC scores, the stories generated by \method are more consistent with the desired psychological state chains, either need or emotion.

\subsubsection{Manual Evaluation}

We perform a manual evaluation between our model and baselines. We randomly generate 100 stories from the test set. For each story, we hire three annotators to give a score in terms of content quality (fluency\&coherence) and content rationality (need\&emotion). For each aspect, we use an average score of the three annotations. We adopt majority voting to make the final decisions among the annotators. As shown in Table \ref{table-humaneval}, all the results show that our model outperforms baselines significantly in fluency, coherence, and psychological state consistency.

\section{Discussion and Analysis}

\subsection{Ablation Study}
An ablation study is conducted on the Story Commonsense dataset to examine the impact of each module separately. We train the model each time by excluding one of our model's modules. And, we summarize the results in Table \ref{table-ablation}.
The results illustrate the harms that the elimination of each of the proposed modules from \method architecture could cause. This attests to the effectiveness of all proposed approaches in the generation of higher controllable stories and subsequently resulting in more accurate evaluation metrics.
As this table demonstrates, the NC/EC accuracy drops the most by ablating the psychological state planner and the psychology controller, which shows that they have the most significant role in composing high-quality psychology-guided stories and consequently accurate evaluation metrics. Besides, psychological state chains (need/emotion chains) all contributed to the high-quality psychology-guided stories.

\subsection{Human A/B Test}
\label{appdix-ab}
We try to compare our model with other baselines by conducting a Human A/B test.
Particularly, we randomly sample 100 examples each for our model and baseline models.
Three annotators are given generated responses from either our model or baselines in random order and are required to choose a better option.
They can either choose one of the responses or select “Tie” when the quality of provided options is hard to access.
Results in Table \ref{table-abtest} confirm that the responses from \method are more preferred by human judges.

\subsection{Case Study}

In this section, we present some generated examples in Table \ref{table-case}.
We select need-related keywords and emotion-related keywords (based on our observation), which are highlighted in corresponding colors, respectively.
From the third block, \method can generate more natural and reasonable psychology-guided stories than baselines.
Since the proposed models can generate stories conditioned on the protagonist’s psychological state chains, they can be used to unfold a story in diverse situations. We demonstrate this capability in the last 4 blocks of Table \ref{table-case} which perform counterfactual transformations on need/emotion chains. 
It shows two examples where for the same leading context, our model can generate stories that follow the counterfactual psychological state chains of the protagonist.

\begin{table}[t]
  \centering
  \scalebox{0.7}{
    \begin{tabular}{ll}
      \toprule[1pt]
      \toprule[1pt]
      \begin{tabular}[c]{@{}l@{}}\textbf{Pgt.}\\ \textbf{Ned.}\\ \textbf{Emo.}\\ \textbf{Cxt.} \end{tabular}   & \begin{tabular}[c]{@{}l@{}} \textit{Tory} \\ \hlb{Stab.} $\rightarrow$ \hly{Love} $\rightarrow$ \sethlcolor{soulorange}\hl{Love} $\rightarrow$ \sethlcolor{soulorange}\hl{Love} $\rightarrow$ \sethlcolor{soulorange}\hl{Love} \\ \sethlcolor{soulyellow}\hl{Fear} $\rightarrow$\sethlcolor{soulyellow}\hl{Fear} $\rightarrow$ \sethlcolor{soulred}\hl{Joy} $\rightarrow$ \sethlcolor{soulred}\hl{Joy} $\rightarrow$\sethlcolor{soulred}\hl{Joy}\\ \textbf{Tory had doubts about getting married.}\end{tabular}\\
      \toprule[0.5pt]
      \textbf{Golden}   & \multicolumn{1}{p{8.5cm}}{She talked to her fiance about their decision. The groom reassured her that he loved her.  She remembered how much she loved him too.  The wedding went forward without anymore problems.} \\
      \toprule[0.5pt]
      \textbf{GPT-2}  & \multicolumn{1}{p{8.5cm}}{She called her husband. She got the answer and \sethlcolor{soulred}\hl{laugh}. She go to shopping then. She bought so many clothes \sethlcolor{soulred}\hl{she like}.}   \\
      \textbf{BART}   & \multicolumn{1}{p{8.5cm}}{She talked to her boyfriend about their relationship. Her boyfriend \sethlcolor{soulorange}\hl{loved her} and gave her a \sethlcolor{soulorange}\hl{kiss}.  She was very \sethlcolor{soulred}\hl{happy} then. They went on the date outside \sethlcolor{soulred}\hl{for fun}.} \\
      \method   & \multicolumn{1}{p{8.5cm}}{Tory is {afraid} about their marriage. Her boyfriend \sethlcolor{soulorange}\hl{gave a gift} as a \sethlcolor{soulred}\hl{blessing}. Tory was \sethlcolor{soulred}\hl{happy} about the gift, and \sethlcolor{soulorange}\hl{reply him with a kiss}. Tory \sethlcolor{soulorange}\hl{went home} to prepare a \sethlcolor{soulred}\hl{surprise} for him.}    \\
      \toprule[1pt]
      \toprule[1pt]
      \textbf{Ned.-CF}   & \hlb{Stab.} $\rightarrow$ \sethlcolor{soulorange}\hl{Love} $\rightarrow$ \sethlcolor{soulorange}\hl{Love} $\rightarrow$\sethlcolor{soulviolet}\hl{Phys.} $\rightarrow$ \sethlcolor{soulviolet}\hl{Phys.} \\
      \toprule[0.5pt]
      \method   & \multicolumn{1}{p{8.5cm}}{She talked to her husband about their marriage.  Her husband \sethlcolor{soulorange}\hl{loved} her and \sethlcolor{soulviolet}\hl{take her to the romantic dinner}.  It is time for them to \sethlcolor{soulviolet}\hl{have a dinner}. She was \sethlcolor{soulred}\hl{happy} with the \sethlcolor{soulviolet}\hl{delicious dinner}.} \\
      \toprule[1pt]
      \toprule[1pt]
      \textbf{Emo.-CF}   & \sethlcolor{soulyellow}\hl{Fear} $\rightarrow$\sethlcolor{soulyellow}\hl{Fear} $\rightarrow$\sethlcolor{soulgreen}\hl{Anger} $\rightarrow$\sethlcolor{soulgreen}\hl{Anger} $\rightarrow$\sethlcolor{soulgreen}\hl{Anger} \\
      \toprule[0.5pt]
      \method   & \multicolumn{1}{p{8.5cm}}{She and her husband talked about their marriage. Her husband was \sethlcolor{soulgreen}\hl{angry} at her doubts. She \sethlcolor{soulgreen}\hl{rushed and fought with} \sethlcolor{soulorange}\hl{his husband}. She \sethlcolor{soulgreen}\hl{gave him a slap} and \sethlcolor{soulorange}\hl{broke up} \sethlcolor{soulgreen}\hl{unhappy}.} \\
      \toprule[1pt]
  \end{tabular}}
  \caption{Generated stories by different models with need and emotion chains. Each psychological state and its corresponding tokens are highlighted in the same color.
    \textbf{CF} represents performing counterfactual transformation on need chains or emotion chains.}
  \label{table-case}
\end{table}

\begin{figure}[t]
  \centering
  \includegraphics[width=0.45\textwidth]{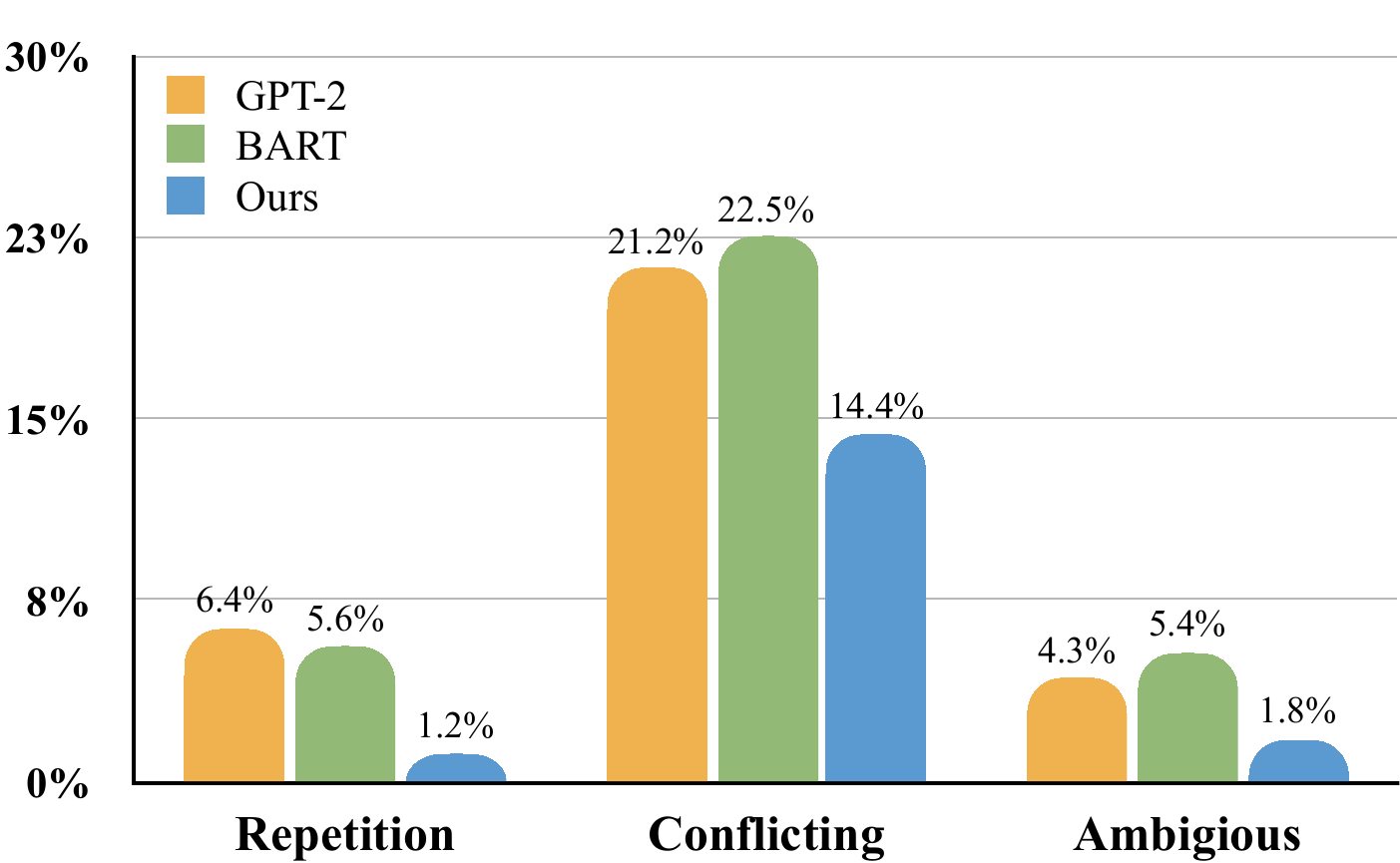} 
  \caption{Distribution of error types for \method (ours) and baseline models (GPT-2 and BART).}
  \label{fig-errorstatic}
\end{figure}

\subsection{Error Analysis}
Although the proposed model outperforms the state-of-the-art baselines, it needs to be noted that there are still many unreasonable stories losing to other models in human evaluation. Therefore, we analyzed error types by manually checking all lost stories in pairwise comparisons between our model and two strong baselines including GPT-2 and BART to reveal the factors that affect the performance. The numbers of stories which lost to our model are 56/64 of 100/100 in total for GPT-2 and BART, respectively. And there are 61 stories of 200 generated by \method losing to these two baselines.

We conclude three main types of error from the lost stories: \textbf{repetition} (repeating the same actions about the need/emotion), \textbf{conflicting psychological state} (wrong causal relation about psychological state), and \textbf{ambiguous psychological state} (difficult to understand the psychological state).
The distribution of different error types is shown in Figure \ref{fig-errorstatic}. We can observe that conflicting and ambiguous psychological state make up most of the errors for all the models.
Compared with GPT-2 and BART, \method reduces chaotic scenes effectively but still suffers from severe repetition, as shown in Table \ref{table-errorcase}. However, the analysis result illustrates that generating a reasonable psychology-guided story is a challenging task.

\begin{table}[t]
  \centering
  \scalebox{0.75}{
    \begin{tabular}{ll}
      \toprule[1pt]
      \toprule[1pt]
      \textbf{Error Type} & \textbf{Cases} \\
      \toprule[0.5pt]
      \textbf{Repetition}     & \multicolumn{1}{p{7.5cm}}{I went to a friends house for a party last weekend. \textit{I was so excited to go.} We played a lot of games that night. \textit{I was so excited to go.} We also had a lot of food to eat.}  \\
      \textbf{Conflicting}     &\multicolumn{1}{p{7.5cm}}{Zack wanted to vote in the election. Unfortunately Zack was traveling during the election. \textit{Zack went to the school for taking class.} Zack then voted again. Zack was happy to have voted.} \\
      \textbf{Ambiguous}     &\multicolumn{1}{p{7.5cm}}{Alice made a cake for her mother. \textit{The cake is so sweet that her mother disliked it. But her mother was very happy and encouraged Alice. Alice was sad because of the failed cooking.} They hugged and smiled in the end.} \\
      \toprule[1pt]
      \toprule[1pt]
  \end{tabular}}
  \caption{Cases of different typical errors. \textit{Italic} words denote the error story events.}
  \label{table-errorcase}
\end{table}

\section{Conclusion and Future Work}
In this paper, we propose a \method system to generate controllable stories that adhere to the story context and protagonist's psychological state chains.
Specifically, we model and integrate local and global psychological states of the protagonist as the story progress.
Experiments demonstrate that \method significantly outperforms baselines and each part shows effectiveness.
In future work, it is important to build a large-scale dataset for developing psychology-guided controllable story generation, regarding aspects of multilingual and long text. Besides, our methodology can be generalized to a wide range of areas, such as automatic storytelling systems and intelligent education agents.

\section*{Acknowledgement}

We thank all anonymous reviewers for their constructive comments and useful advice. Also thanks for the discussion with Yajing Sun, Yongxiu Xu, and Ping Guo. This work is supported by the National Natural Science Foundation of China (No.62006222 and No.U21B2009).
Thanks for COLING organizers and the proposed pre-trained language models, data, codes. 

\noindent\textbf{Contribution List} Yuqiang Xie: Idea, Paper Writing, Coding; Yue Hu: Guiding, Discussion; Yunpeng Li: Coding; Guanqun Bi: Paper Polish, Discussion; Luxi Xing: Review; Wei Peng: Review.

Thanks for the hard work and dedication of all team members. 

\bibliography{acl_latex}

\begin{thebibliography}{57}
\expandafter\ifx\csname natexlab\endcsname\relax\def\natexlab#1{#1}\fi

\bibitem[{Alabdulkarim et~al.(2021)Alabdulkarim, Li, and
  Peng}]{Alabdulkarim2021AutomaticSG}
Amal Alabdulkarim, Siyan Li, and Xiangyu Peng. 2021.
\newblock \href {https://doi.org/10.18653/v1/2021.nuse-1.8} {Automatic story
  generation: Challenges and attempts}.
\newblock In \emph{Proceedings of the Third Workshop on Narrative
  Understanding}, pages 72--83, Virtual. Association for Computational
  Linguistics.

\bibitem[{Ammanabrolu et~al.(2021)Ammanabrolu, Cheung, Broniec, and
  Riedl}]{Ammanabrolu2021AutomatedSV}
Prithviraj Ammanabrolu, W.~Cheung, William Broniec, and Mark~O. Riedl. 2021.
\newblock Automated storytelling via causal, commonsense plot ordering.
\newblock In \emph{AAAI}.

\bibitem[{Bosselut et~al.(2019)Bosselut, Rashkin, Sap, Malaviya, Celikyilmaz,
  and Choi}]{DBLP:conf/acl/BosselutRSMCC19}
Antoine Bosselut, Hannah Rashkin, Maarten Sap, Chaitanya Malaviya, Asli
  Celikyilmaz, and Yejin Choi. 2019.
\newblock \href {https://doi.org/10.18653/v1/P19-1470} {{COMET}: Commonsense
  transformers for automatic knowledge graph construction}.
\newblock In \emph{Proceedings of the 57th Annual Meeting of the Association
  for Computational Linguistics}, pages 4762--4779, Florence, Italy.
  Association for Computational Linguistics.

\bibitem[{Brahman and Chaturvedi(2020)}]{DBLP:conf/emnlp/BrahmanC20}
Faeze Brahman and Snigdha Chaturvedi. 2020.
\newblock \href {https://doi.org/10.18653/v1/2020.emnlp-main.426} {Modeling
  protagonist emotions for emotion-aware storytelling}.
\newblock In \emph{Proceedings of the 2020 Conference on Empirical Methods in
  Natural Language Processing (EMNLP)}, pages 5277--5294, Online. Association
  for Computational Linguistics.

\bibitem[{Budzianowski and Vuli{\'c}(2019)}]{Budzianowski2019HelloIG}
Pawe{\l} Budzianowski and Ivan Vuli{\'c}. 2019.
\newblock \href {https://doi.org/10.18653/v1/D19-5602} {Hello, it{'}s {GPT}-2 -
  how can {I} help you? towards the use of pretrained language models for
  task-oriented dialogue systems}.
\newblock In \emph{Proceedings of the 3rd Workshop on Neural Generation and
  Translation}, pages 15--22, Hong Kong. Association for Computational
  Linguistics.

\bibitem[{Cho et~al.(2014)Cho, van Merri{\"e}nboer, Bahdanau, and
  Bengio}]{Cho2014OnTP}
Kyunghyun Cho, Bart van Merri{\"e}nboer, Dzmitry Bahdanau, and Yoshua Bengio.
  2014.
\newblock \href {https://doi.org/10.3115/v1/W14-4012} {On the properties of
  neural machine translation: Encoder{--}decoder approaches}.
\newblock In \emph{Proceedings of {SSST}-8, Eighth Workshop on Syntax,
  Semantics and Structure in Statistical Translation}, pages 103--111, Doha,
  Qatar. Association for Computational Linguistics.

\bibitem[{Clark et~al.(2018)Clark, Ji, and Smith}]{DBLP:conf/naacl/ClarkJS18}
Elizabeth Clark, Yangfeng Ji, and Noah~A. Smith. 2018.
\newblock \href {https://doi.org/10.18653/v1/N18-1204} {Neural text generation
  in stories using entity representations as context}.
\newblock In \emph{Proceedings of the 2018 Conference of the North {A}merican
  Chapter of the Association for Computational Linguistics: Human Language
  Technologies, Volume 1 (Long Papers)}, pages 2250--2260, New Orleans,
  Louisiana. Association for Computational Linguistics.

\bibitem[{Dathathri et~al.(2020)Dathathri, Madotto, Lan, Hung, Frank, Molino,
  Yosinski, and Liu}]{Dathathri2020PlugAP}
Sumanth Dathathri, Andrea Madotto, Janice Lan, Jane Hung, Eric Frank, Piero
  Molino, Jason Yosinski, and Rosanne Liu. 2020.
\newblock \href {https://openreview.net/forum?id=H1edEyBKDS} {Plug and play
  language models: {A} simple approach to controlled text generation}.
\newblock In \emph{8th International Conference on Learning Representations,
  {ICLR} 2020, Addis Ababa, Ethiopia, April 26-30, 2020}. OpenReview.net.

\bibitem[{Fan et~al.(2018{\natexlab{a}})Fan, Grangier, and
  Auli}]{Fan2018ControllableAS}
Angela Fan, David Grangier, and Michael Auli. 2018{\natexlab{a}}.
\newblock \href {https://doi.org/10.18653/v1/W18-2706} {Controllable
  abstractive summarization}.
\newblock In \emph{Proceedings of the 2nd Workshop on Neural Machine
  Translation and Generation}, pages 45--54, Melbourne, Australia. Association
  for Computational Linguistics.

\bibitem[{Fan et~al.(2018{\natexlab{b}})Fan, Lewis, and
  Dauphin}]{Fan2018HierarchicalNS}
Angela Fan, Mike Lewis, and Yann Dauphin. 2018{\natexlab{b}}.
\newblock \href {https://doi.org/10.18653/v1/P18-1082} {Hierarchical neural
  story generation}.
\newblock In \emph{Proceedings of the 56th Annual Meeting of the Association
  for Computational Linguistics (Volume 1: Long Papers)}, pages 889--898,
  Melbourne, Australia. Association for Computational Linguistics.

\bibitem[{Fan et~al.(2019)Fan, Lewis, and Dauphin}]{Fan2019StrategiesFS}
Angela Fan, Mike Lewis, and Yann Dauphin. 2019.
\newblock \href {https://doi.org/10.18653/v1/P19-1254} {Strategies for
  structuring story generation}.
\newblock In \emph{Proceedings of the 57th Annual Meeting of the Association
  for Computational Linguistics}, pages 2650--2660, Florence, Italy.
  Association for Computational Linguistics.

\bibitem[{Guan et~al.(2020)Guan, Huang, Zhao, Zhu, and Huang}]{Guan2020AKP}
Jian Guan, Fei Huang, Zhihao Zhao, Xiaoyan Zhu, and Minlie Huang. 2020.
\newblock \href {https://doi.org/10.1162/tacl_a_00302} {A knowledge-enhanced
  pretraining model for commonsense story generation}.
\newblock \emph{Transactions of the Association for Computational Linguistics},
  8:93--108.

\bibitem[{Huang et~al.(2018)Huang, Za{\"\i}ane, Trabelsi, and
  Dziri}]{Huang2018AutomaticDG}
Chenyang Huang, Osmar Za{\"\i}ane, Amine Trabelsi, and Nouha Dziri. 2018.
\newblock \href {https://doi.org/10.18653/v1/N18-2008} {Automatic dialogue
  generation with expressed emotions}.
\newblock In \emph{Proceedings of the 2018 Conference of the North {A}merican
  Chapter of the Association for Computational Linguistics: Human Language
  Technologies, Volume 2 (Short Papers)}, pages 49--54, New Orleans, Louisiana.
  Association for Computational Linguistics.

\bibitem[{Jain et~al.(2017)Jain, Agrawal, Mishra, Sukhwani, Laha, and
  Sankaranarayanan}]{Jain2017StoryGF}
Parag Jain, Priyanka Agrawal, Abhijit Mishra, Mohak Sukhwani, Anirban Laha, and
  Karthik Sankaranarayanan. 2017.
\newblock Story generation from sequence of independent short descriptions.
\newblock \emph{ArXiv}, abs/1707.05501.

\bibitem[{Kikuchi et~al.(2016)Kikuchi, Neubig, Sasano, Takamura, and
  Okumura}]{Kikuchi2016ControllingOL}
Yuta Kikuchi, Graham Neubig, Ryohei Sasano, Hiroya Takamura, and Manabu
  Okumura. 2016.
\newblock \href {https://doi.org/10.18653/v1/D16-1140} {Controlling output
  length in neural encoder-decoders}.
\newblock In \emph{Proceedings of the 2016 Conference on Empirical Methods in
  Natural Language Processing}, pages 1328--1338, Austin, Texas. Association
  for Computational Linguistics.

\bibitem[{Kong et~al.(2021)Kong, Huang, Tung, Guan, and
  Huang}]{Kong2021StylizedSG}
Xiangzhe Kong, Jialiang Huang, Ziquan Tung, Jian Guan, and Minlie Huang. 2021.
\newblock \href {https://doi.org/10.18653/v1/2021.findings-acl.215} {Stylized
  story generation with style-guided planning}.
\newblock In \emph{Findings of the Association for Computational Linguistics:
  ACL-IJCNLP 2021}, pages 2430--2436, Online. Association for Computational
  Linguistics.

\bibitem[{Lewis et~al.(2020)Lewis, Liu, Goyal, Ghazvininejad, Mohamed, Levy,
  Stoyanov, and Zettlemoyer}]{DBLP:conf/acl/LewisLGGMLSZ20}
Mike Lewis, Yinhan Liu, Naman Goyal, Marjan Ghazvininejad, Abdelrahman Mohamed,
  Omer Levy, Veselin Stoyanov, and Luke Zettlemoyer. 2020.
\newblock \href {https://doi.org/10.18653/v1/2020.acl-main.703} {{BART}:
  Denoising sequence-to-sequence pre-training for natural language generation,
  translation, and comprehension}.
\newblock In \emph{Proceedings of the 58th Annual Meeting of the Association
  for Computational Linguistics}, pages 7871--7880, Online. Association for
  Computational Linguistics.

\bibitem[{Lin(2004)}]{Lin2004ROUGEAP}
Chin-Yew Lin. 2004.
\newblock \href {https://aclanthology.org/W04-1013} {{ROUGE}: A package for
  automatic evaluation of summaries}.
\newblock In \emph{Text Summarization Branches Out}, pages 74--81, Barcelona,
  Spain. Association for Computational Linguistics.

\bibitem[{Liu et~al.(2020)Liu, Li, Yu, Huang, Liu, Zhao, and
  Yan}]{DBLP:conf/aaai/LiuLYHL0020}
Danyang Liu, Juntao Li, Meng{-}Hsuan Yu, Ziming Huang, Gongshen Liu, Dongyan
  Zhao, and Rui Yan. 2020.
\newblock \href {https://aaai.org/ojs/index.php/AAAI/article/view/5536} {A
  character-centric neural model for automated story generation}.
\newblock In \emph{The Thirty-Fourth {AAAI} Conference on Artificial
  Intelligence, {AAAI} 2020}, pages 1725--1732. {AAAI} Press.

\bibitem[{Liu et~al.(2019)Liu, Ott, Goyal, Du, Joshi, Chen, Levy, Lewis,
  Zettlemoyer, and Stoyanov}]{Liu2019RoBERTaAR}
Yinhan Liu, Myle Ott, Naman Goyal, Jingfei Du, Mandar Joshi, Danqi Chen, Omer
  Levy, M.~Lewis, Luke Zettlemoyer, and Veselin Stoyanov. 2019.
\newblock Roberta: A robustly optimized bert pretraining approach.
\newblock \emph{ArXiv}, abs/1907.11692.

\bibitem[{Luo et~al.(2019)Luo, Dai, Yang, Liu, Chang, Sui, and
  Sun}]{DBLP:conf/acl/LuoDYLCSS19}
Fuli Luo, Damai Dai, Pengcheng Yang, Tianyu Liu, Baobao Chang, Zhifang Sui, and
  Xu~Sun. 2019.
\newblock \href {https://doi.org/10.18653/v1/P19-1603} {Learning to control the
  fine-grained sentiment for story ending generation}.
\newblock In \emph{Proceedings of the 57th Annual Meeting of the Association
  for Computational Linguistics}, pages 6020--6026, Florence, Italy.
  Association for Computational Linguistics.

\bibitem[{Martin et~al.(2018)Martin, Ammanabrolu, Wang, Hancock, Singh,
  Harrison, and Riedl}]{Martin2018EventRF}
Lara~J. Martin, Prithviraj Ammanabrolu, Xinyu Wang, William Hancock, Shruti
  Singh, Brent Harrison, and Mark~O. Riedl. 2018.
\newblock \href
  {https://www.aaai.org/ocs/index.php/AAAI/AAAI18/paper/view/17046} {Event
  representations for automated story generation with deep neural nets}.
\newblock In \emph{Proceedings of the Thirty-Second {AAAI} Conference on
  Artificial Intelligence, (AAAI-18)}, pages 868--875. {AAAI} Press.

\bibitem[{Maslow(1943)}]{Maslow2013ATO}
Abraham~Harold Maslow. 1943.
\newblock A theory of human motivation.
\newblock In \emph{Psychological review}.

\bibitem[{Morrow(1985)}]{Morrow1985ProminentCA}
D.~Morrow. 1985.
\newblock Prominent characters and events organize narrative understanding.
\newblock \emph{Journal of Memory and Language}, 24:304--319.

\bibitem[{Papineni et~al.(2002)Papineni, Roukos, Ward, and
  Zhu}]{DBLP:conf/acl/PapineniRWZ02}
Kishore Papineni, Salim Roukos, Todd Ward, and Wei-Jing Zhu. 2002.
\newblock \href {https://doi.org/10.3115/1073083.1073135} {{B}leu: a method for
  automatic evaluation of machine translation}.
\newblock In \emph{Proceedings of the 40th Annual Meeting of the Association
  for Computational Linguistics}, pages 311--318, Philadelphia, Pennsylvania,
  USA. Association for Computational Linguistics.

\bibitem[{Paszke et~al.(2019)Paszke, Gross, Massa, Lerer, Bradbury, Chanan,
  Killeen, Lin, Gimelshein, Antiga, Desmaison, K{\"{o}}pf, Yang, DeVito,
  Raison, Tejani, Chilamkurthy, Steiner, Fang, Bai, and
  Chintala}]{DBLP:conf/nips/PaszkeGMLBCKLGA19}
Adam Paszke, Sam Gross, Francisco Massa, Adam Lerer, James Bradbury, Gregory
  Chanan, Trevor Killeen, Zeming Lin, Natalia Gimelshein, Luca Antiga, Alban
  Desmaison, Andreas K{\"{o}}pf, Edward Yang, Zachary DeVito, Martin Raison,
  Alykhan Tejani, Sasank Chilamkurthy, Benoit Steiner, Lu~Fang, Junjie Bai, and
  Soumith Chintala. 2019.
\newblock \href
  {https://proceedings.neurips.cc/paper/2019/hash/bdbca288fee7f92f2bfa9f7012727740-Abstract.html}
  {Pytorch: An imperative style, high-performance deep learning library}.
\newblock In \emph{Advances in Neural Information Processing Systems 32: Annual
  Conference on Neural Information Processing Systems 2019, NeurIPS 2019,
  December 8-14, 2019, Vancouver, BC, Canada}, pages 8024--8035.

\bibitem[{Paul and Frank(2021)}]{Paul2021COINSDG}
Debjit Paul and Anette Frank. 2021.
\newblock \href {https://doi.org/10.18653/v1/2021.acl-long.395} {{COINS}:
  Dynamically generating {CO}ntextualized inference rules for narrative story
  completion}.
\newblock In \emph{Proceedings of the 59th Annual Meeting of the Association
  for Computational Linguistics and the 11th International Joint Conference on
  Natural Language Processing (Volume 1: Long Papers)}, pages 5086--5099,
  Online. Association for Computational Linguistics.

\bibitem[{Peng et~al.(2018)Peng, Ghazvininejad, May, and
  Knight}]{Peng2018TowardsCS}
Nanyun Peng, Marjan Ghazvininejad, Jonathan May, and Kevin Knight. 2018.
\newblock \href {https://doi.org/10.18653/v1/W18-1505} {Towards controllable
  story generation}.
\newblock In \emph{Proceedings of the First Workshop on Storytelling}, pages
  43--49, New Orleans, Louisiana. Association for Computational Linguistics.

\bibitem[{Peng et~al.(2022)Peng, Hu, Xing, Xie, Sun, and Li}]{ijcai2022wei}
Wei Peng, Yue Hu, Luxi Xing, Yuqiang Xie, Yajing Sun, and Yunpeng Li. 2022.
\newblock \href {https://doi.org/10.24963/ijcai.2022/600} {Control globally,
  understand locally: {A} global-to-local hierarchical graph network for
  emotional support conversation}.
\newblock In \emph{Proceedings of the Thirty-First International Joint
  Conference on Artificial Intelligence, {IJCAI} 2022, Vienna, Austria, 23-29
  July 2022}, pages 4324--4330. ijcai.org.

\bibitem[{Pennington et~al.(2014)Pennington, Socher, and
  Manning}]{Pennington2014GloVeGV}
Jeffrey Pennington, Richard Socher, and Christopher Manning. 2014.
\newblock \href {https://doi.org/10.3115/v1/D14-1162} {{G}lo{V}e: Global
  vectors for word representation}.
\newblock In \emph{Proceedings of the 2014 Conference on Empirical Methods in
  Natural Language Processing ({EMNLP})}, pages 1532--1543, Doha, Qatar.
  Association for Computational Linguistics.

\bibitem[{P{\'e}rez and Sharples(2001)}]{Prez2001MEXICAAC}
R.~Y. P{\'e}rez and M.~Sharples. 2001.
\newblock Mexica: A computer model of a cognitive account of creative writing.
\newblock \emph{Journal of Experimental \& Theoretical Artificial
  Intelligence}, 13:119 -- 139.

\bibitem[{Plutchik(1980)}]{Plutchik1980AGP}
Robert Plutchik. 1980.
\newblock A general psychoevolutionary theory of emotion.
\newblock In \emph{Theories of emotion}.

\bibitem[{Porteous and Cavazza(2009)}]{Porteous2009ControllingNG}
J.~Porteous and M.~Cavazza. 2009.
\newblock Controlling narrative generation with planning trajectories: The role
  of constraints.
\newblock In \emph{ICIDS}.

\bibitem[{Qin et~al.(2019)Qin, Bosselut, Holtzman, Bhagavatula, Clark, and
  Choi}]{Qin2019CounterfactualSR}
Lianhui Qin, Antoine Bosselut, Ari Holtzman, Chandra Bhagavatula, Elizabeth
  Clark, and Yejin Choi. 2019.
\newblock \href {https://doi.org/10.18653/v1/D19-1509} {Counterfactual story
  reasoning and generation}.
\newblock In \emph{Proceedings of the 2019 Conference on Empirical Methods in
  Natural Language Processing and the 9th International Joint Conference on
  Natural Language Processing (EMNLP-IJCNLP)}, pages 5043--5053, Hong Kong,
  China. Association for Computational Linguistics.

\bibitem[{Radford et~al.(2019)Radford, Wu, Child, Luan, Amodei, and
  Sutskever}]{Radford2019LanguageMA}
Alec Radford, Jeff Wu, Rewon Child, David Luan, Dario Amodei, and Ilya
  Sutskever. 2019.
\newblock Language models are unsupervised multitask learners.
\newblock In \emph{OpenAI Blog}.

\bibitem[{Rashkin et~al.(2018)Rashkin, Bosselut, Sap, Knight, and
  Choi}]{DBLP:conf/acl/KnightCSRB18}
Hannah Rashkin, Antoine Bosselut, Maarten Sap, Kevin Knight, and Yejin Choi.
  2018.
\newblock \href {https://doi.org/10.18653/v1/P18-1213} {Modeling naive
  psychology of characters in simple commonsense stories}.
\newblock In \emph{Proceedings of the 56th Annual Meeting of the Association
  for Computational Linguistics (Volume 1: Long Papers)}, pages 2289--2299,
  Melbourne, Australia. Association for Computational Linguistics.

\bibitem[{Rashkin et~al.(2020)Rashkin, Celikyilmaz, Choi, and
  Gao}]{Rashkin2020PlotMachinesOG}
Hannah Rashkin, Asli Celikyilmaz, Yejin Choi, and Jianfeng Gao. 2020.
\newblock \href {https://doi.org/10.18653/v1/2020.emnlp-main.349}
  {{P}lot{M}achines: Outline-conditioned generation with dynamic plot state
  tracking}.
\newblock In \emph{Proceedings of the 2020 Conference on Empirical Methods in
  Natural Language Processing (EMNLP)}, pages 4274--4295, Online. Association
  for Computational Linguistics.

\bibitem[{Ricoeur(1984)}]{ricoeur1984time}
Paul Ricoeur. 1984.
\newblock Time and narrative.

\bibitem[{Riedl and Young(2010)}]{Riedl2010NarrativePB}
Mark~O. Riedl and R.~M. Young. 2010.
\newblock Narrative planning: Balancing plot and character.
\newblock \emph{ArXiv}, abs/1401.3841.

\bibitem[{Roemmele(2016)}]{Roemmele2016WritingSW}
Melissa Roemmele. 2016.
\newblock \href
  {http://www.aaai.org/ocs/index.php/AAAI/AAAI16/paper/view/11966} {Writing
  stories with help from recurrent neural networks}.
\newblock In \emph{Proceedings of the Thirtieth {AAAI} Conference on Artificial
  Intelligence, February 12-17, 2016, Phoenix, Arizona, {USA}}, pages
  4311--4342. {AAAI} Press.

\bibitem[{Song et~al.(2019)Song, Zheng, Liu, Xu, and
  Huang}]{DBLP:conf/acl/SongZLXH19}
Zhenqiao Song, Xiaoqing Zheng, Lu~Liu, Mu~Xu, and Xuanjing Huang. 2019.
\newblock \href {https://doi.org/10.18653/v1/P19-1359} {Generating responses
  with a specific emotion in dialog}.
\newblock In \emph{Proceedings of the 57th Annual Meeting of the Association
  for Computational Linguistics}, pages 3685--3695, Florence, Italy.
  Association for Computational Linguistics.

\bibitem[{Tambwekar et~al.(2019)Tambwekar, Dhuliawala, Martin, Mehta, Harrison,
  and Riedl}]{Tambwekar2019ControllableNS}
Pradyumna Tambwekar, Murtaza Dhuliawala, Lara~J. Martin, Animesh Mehta, Brent
  Harrison, and Mark~O. Riedl. 2019.
\newblock \href {https://doi.org/10.24963/ijcai.2019/829} {Controllable neural
  story plot generation via reward shaping}.
\newblock In \emph{Proceedings of the Twenty-Eighth International Joint
  Conference on Artificial Intelligence, {IJCAI} 2019, Macao, China, August
  10-16, 2019}, pages 5982--5988. ijcai.org.

\bibitem[{Vaswani et~al.(2017)Vaswani, Shazeer, Parmar, Uszkoreit, Jones,
  Gomez, Kaiser, and Polosukhin}]{DBLP:conf/nips/VaswaniSPUJGKP17}
Ashish Vaswani, Noam Shazeer, Niki Parmar, Jakob Uszkoreit, Llion Jones,
  Aidan~N. Gomez, Lukasz Kaiser, and Illia Polosukhin. 2017.
\newblock \href
  {https://proceedings.neurips.cc/paper/2017/hash/3f5ee243547dee91fbd053c1c4a845aa-Abstract.html}
  {Attention is all you need}.
\newblock In \emph{Advances in Neural Information Processing Systems 30: Annual
  Conference on Neural Information Processing Systems 2017, December 4-9, 2017,
  Long Beach, CA, {USA}}, pages 5998--6008.

\bibitem[{Vonnegut(1981)}]{Vonnegut1981PalmSA}
K.~Vonnegut. 1981.
\newblock Palm sunday: An autobiographical collage.
\newblock In \emph{RosetTaBooks, LLC New York}.

\bibitem[{Wang et~al.(2017)Wang, Jojic, Brockett, and
  Nyberg}]{Wang2017SteeringOS}
Di~Wang, Nebojsa Jojic, Chris Brockett, and Eric Nyberg. 2017.
\newblock \href {https://doi.org/10.18653/v1/D17-1228} {Steering output style
  and topic in neural response generation}.
\newblock In \emph{Proceedings of the 2017 Conference on Empirical Methods in
  Natural Language Processing}, pages 2140--2150, Copenhagen, Denmark.
  Association for Computational Linguistics.

\bibitem[{Weber et~al.(2020)Weber, Shekhar, Kwon, Balasubramanian, and
  Chambers}]{DBLP:conf/conll/WeberSKBC20}
Noah Weber, Leena Shekhar, Heeyoung Kwon, Niranjan Balasubramanian, and
  Nathanael Chambers. 2020.
\newblock \href {https://doi.org/10.18653/v1/2020.conll-1.42} {Generating
  narrative text in a switching dynamical system}.
\newblock In \emph{Proceedings of the 24th Conference on Computational Natural
  Language Learning}, pages 520--530, Online. Association for Computational
  Linguistics.

\bibitem[{Wolf et~al.(2020)Wolf, Debut, Sanh, Chaumond, Delangue, Moi, Cistac,
  Rault, Louf, Funtowicz, Davison, Shleifer, von Platen, Ma, Jernite, Plu, Xu,
  Le~Scao, Gugger, Drame, Lhoest, and Rush}]{DBLP:conf/emnlp/WolfDSCDMCRLFDS20}
Thomas Wolf, Lysandre Debut, Victor Sanh, Julien Chaumond, Clement Delangue,
  Anthony Moi, Pierric Cistac, Tim Rault, Remi Louf, Morgan Funtowicz, Joe
  Davison, Sam Shleifer, Patrick von Platen, Clara Ma, Yacine Jernite, Julien
  Plu, Canwen Xu, Teven Le~Scao, Sylvain Gugger, Mariama Drame, Quentin Lhoest,
  and Alexander Rush. 2020.
\newblock \href {https://doi.org/10.18653/v1/2020.emnlp-demos.6} {Transformers:
  State-of-the-art natural language processing}.
\newblock In \emph{Proceedings of the 2020 Conference on Empirical Methods in
  Natural Language Processing: System Demonstrations}, pages 38--45, Online.
  Association for Computational Linguistics.

\bibitem[{Wolf et~al.(2019)Wolf, Sanh, Chaumond, and
  Delangue}]{Wolf2019TransferTransfoAT}
Thomas Wolf, Victor Sanh, Julien Chaumond, and Clement Delangue. 2019.
\newblock Transfertransfo: A transfer learning approach for neural network
  based conversational agents.
\newblock \emph{ArXiv}, abs/1901.08149.

\bibitem[{Xie et~al.(2022{\natexlab{a}})Xie, Hu, Peng, Bi, and
  Xing}]{xie2022comma}
Yuqiang Xie, Yue Hu, Wei Peng, Guanqun Bi, and Luxi Xing. 2022{\natexlab{a}}.
\newblock \href {https://doi.org/10.48550/ARXIV.2209.06470} {Comma: Modeling
  relationship among motivations, emotions and actions in language-based human
  activities}.

\bibitem[{Xie et~al.(2022{\natexlab{b}})Xie, Hu, Xing, Li, Peng, and
  Guo}]{clseg2022xie}
Yuqiang Xie, Yue Hu, Luxi Xing, Yunpeng Li, Wei Peng, and Ping Guo.
  2022{\natexlab{b}}.
\newblock \href {https://doi.org/10.1109/ICASSP43922.2022.9747435} {Clseg:
  Contrastive learning of story ending generation}.
\newblock In \emph{ICASSP 2022 - 2022 IEEE International Conference on
  Acoustics, Speech and Signal Processing (ICASSP)}, pages 8057--8061.

\bibitem[{Xu et~al.(2020{\natexlab{a}})Xu, Wang, Ma, Tresp, Wang, Zhou, and
  Du}]{Xu2020ControllableMP}
Feifei Xu, Xinpeng Wang, Yunpu Ma, Volker Tresp, Yuyi Wang, Shanlin Zhou, and
  Haizhou Du. 2020{\natexlab{a}}.
\newblock \href {https://doi.org/10.1145/3340531.3411937} {Controllable
  multi-character psychology-oriented story generation}.
\newblock In \emph{{CIKM} '20: The 29th {ACM} International Conference on
  Information and Knowledge Management, Virtual Event, Ireland, October 19-23,
  2020}, pages 1675--1684. {ACM}.

\bibitem[{Xu et~al.(2018)Xu, Ren, Zhang, Zeng, Cai, and Sun}]{Xu2018ASM}
Jingjing Xu, Xuancheng Ren, Yi~Zhang, Qi~Zeng, Xiaoyan Cai, and Xu~Sun. 2018.
\newblock \href {https://doi.org/10.18653/v1/D18-1462} {A skeleton-based model
  for promoting coherence among sentences in narrative story generation}.
\newblock In \emph{Proceedings of the 2018 Conference on Empirical Methods in
  Natural Language Processing}, pages 4306--4315, Brussels, Belgium.
  Association for Computational Linguistics.

\bibitem[{Xu et~al.(2020{\natexlab{b}})Xu, Patwary, Shoeybi, Puri, Fung,
  Anandkumar, and Catanzaro}]{xu-etal-2020-megatron}
Peng Xu, Mostofa Patwary, Mohammad Shoeybi, Raul Puri, Pascale Fung, Anima
  Anandkumar, and Bryan Catanzaro. 2020{\natexlab{b}}.
\newblock \href {https://doi.org/10.18653/v1/2020.emnlp-main.226}
  {{MEGATRON}-{CNTRL}: Controllable story generation with external knowledge
  using large-scale language models}.
\newblock In \emph{Proceedings of the 2020 Conference on Empirical Methods in
  Natural Language Processing (EMNLP)}, pages 2831--2845, Online. Association
  for Computational Linguistics.

\bibitem[{Yao et~al.(2019)Yao, Peng, Weischedel, Knight, Zhao, and
  Yan}]{Yao2019PlanAndWriteTB}
Lili Yao, Nanyun Peng, R.~Weischedel, Kevin Knight, Dongyan Zhao, and Rui Yan.
  2019.
\newblock Plan-and-write: Towards better automatic storytelling.
\newblock In \emph{AAAI}.

\bibitem[{Zhang et~al.(2022)Zhang, Yang, Meng, Chen, and
  Zhou}]{DBLP:journals/corr/abs-2203-00819}
Duzhen Zhang, Zhen Yang, Fandong Meng, Xiuyi Chen, and Jie Zhou. 2022.
\newblock \href {https://doi.org/10.48550/arXiv.2203.00819} {{TSAM:} {A}
  two-stream attention model for causal emotion entailment}.
\newblock \emph{CoRR}, abs/2203.00819.

\bibitem[{Zhou et~al.(2018)Zhou, Huang, Zhang, Zhu, and
  Liu}]{Zhou2018EmotionalCM}
Hao Zhou, Minlie Huang, Tianyang Zhang, Xiaoyan Zhu, and Bing Liu. 2018.
\newblock \href
  {https://www.aaai.org/ocs/index.php/AAAI/AAAI18/paper/view/16455} {Emotional
  chatting machine: Emotional conversation generation with internal and
  external memory}.
\newblock In \emph{Proceedings of the Thirty-Second {AAAI} Conference on
  Artificial Intelligence, (AAAI-18)}, pages 730--739. {AAAI} Press.

\bibitem[{Zhou and Wang(2018)}]{Zhou2018MojiTalkGE}
Xianda Zhou and William~Yang Wang. 2018.
\newblock \href {https://doi.org/10.18653/v1/P18-1104} {{M}oji{T}alk:
  Generating emotional responses at scale}.
\newblock In \emph{Proceedings of the 56th Annual Meeting of the Association
  for Computational Linguistics (Volume 1: Long Papers)}, pages 1128--1137,
  Melbourne, Australia. Association for Computational Linguistics.

\end{thebibliography}
\bibliographystyle{acl_natbib}

\end{document}